\documentclass[letterpaper]{article}
\usepackage{aaai19}
\usepackage{times}
\usepackage{helvet}
\usepackage{courier}
\usepackage{url}
\usepackage{graphicx}
\usepackage{diagbox}
\usepackage{amsmath}
\usepackage{amssymb}
\usepackage{amsthm}
\usepackage{mathdots}
\usepackage{caption}
\usepackage{subcaption}
\usepackage{color}
\usepackage{subcaption}

\usepackage{mathtools}
\allowdisplaybreaks
\begin{document}

\pdfinfo{
/Title (Evolutionary Game-Theoretical Analysis for General Multiplayer Asymmetric Games)
/Author (Paper 7136)
/Keywords (Input your keywords in this optional area)
}
%
\setcounter{secnumdepth}{2}

%
%
 \author{Xinyu Zhang,\textsuperscript{1}
Peng Peng,\textsuperscript{2}
Yushan Zhou,\textsuperscript{3}
Haifeng Wang,\textsuperscript{4}
Wenxin Li\textsuperscript{3}\\
\textsuperscript{1}{The George Washington University}\\
\textsuperscript{2}{inspir.ai}\\
\textsuperscript{3}{Peking University}\\
\textsuperscript{4}{University College London}\\
zxyhxz@gwu.edu,
pp@inspirai.com, 
zysls@pku.edu.cn,
pkuzhf@pku.edu.cn,
lwx@pku.edu.cn}
\title{Evolutionary Game-Theoretical Analysis for General \\ Multiplayer Asymmetric Games}
\maketitle

\begin{abstract}
Evolutionary game theory has been a successful tool to combine classical game theory with learning-dynamical descriptions in multiagent systems. Provided some symmetric structures of interacting players, many studies have been focused on using a simplified heuristic payoff table as input to analyse the dynamics of interactions. Nevertheless, even for the state-of-the-art method, there are two limits. First, there is inaccuracy when analysing the simplified payoff table. Second, no existing work is able to deal with 2-population multiplayer asymmetric games. In this paper, we fill the gap between heuristic payoff table and dynamic analysis without any inaccuracy. In addition, we propose a general framework for $m$ versus $n$ 2-population multiplayer asymmetric games. Then, we compare our method with the state-of-the-art in some classic games. Finally, to illustrate our method, we perform empirical game-theoretical analysis on Wolfpack as well as StarCraft II, both of which involve complex multiagent interactions.
\end{abstract}

\section{Introduction}
\label{sec:intro}
Evolutionary game theory (EGT) has been successfully applied to analyse the learning dynamics in multiagent systems \cite{tuyls2004evolutionary,tuyls2006evolutionary}. The EGT relaxes the hyper-rationality assumption in traditional game theory \cite{morgenstern1944theory,gibbons1992game} and replaces it with biological mechanism such as natural selection and mutation \cite{smith1973logic,smith1982evolution,weibull1997evolutionary,hofbauer1998evolutionary}. The replicator dynamics (RD), a system of ordinary differential equations, is the core mathematical model in EGT and describes the evolutionary process for the population change of different types over time \cite{taylor1978evolutionary}.
However, using game theory to study multiagent interactions in complex systems is non-trivial. Empirical game-theoretical analysis (EGTA) tries to investigate and analyse meta games abstracted from the original full game \cite{walsh2002analyzing,wellman2006methods}. Instead of trying to detail the decision-making processes at the level of the atomic actions, EGTA focuses on meta-strategies which are discovered and meta-reasoned in the strategy space heuristically. In this way, an abstracted game with much smaller structure is extracted from the original full game. Tuyls et al. provide theoretical guarantees on the approximation of original full game by an estimated game based on sampled data \cite{tuyls2018generalised}. With the higher-level strategies being the primitive actions of the game, one should get a payoff table in the normal form, which can be investigated via EGT in a standard way. Nevertheless, when more than two players interact with each other in the game, the payoff table becomes a higher-order tensor rather than a matrix. This is computationally unacceptable when the number of players is large, since the payoff table would be exponentially larger to the amount of players. Fortunately, if there exists some symmetricities in the payoff structure, which is ubiquitous in most scenarios, the heuristic payoff table (HPT) is able to crucially simplify the payoff table\cite{walsh2002analyzing}. In a nutshell, the HPT makes it amenable to theoretically analyse the multiplayer game as long as the game is symmetric. This method has been applied to many areas such as continuous double auctions, poker games and multi-robot systems \cite{hennes2013evolutionary}.

Recent studies have generalised the HPT-based method for asymmetric games \cite{tuyls2018symmetric,tuyls2018generalised}.
In particular, the state-of-the-art method \cite{tuyls2018generalised} performs a decomposition of the HPT of an asymmetric game into two symmetric counterparts to make it compatible with a previous work \cite{tuyls2018symmetric}, in which the authors link the asymmetric game and the symmetric counterparts by analysing the Nash equilibria. In that way, asymmetric problem is converted to symmetric problems so the formulas for symmetric HPT can be directly utilised. However, there are two limitations. Firstly, there is inaccuracy when analysing the HPT using the method, i.e., there is loss of information, which should not have been the case in the spirit of HPT. Secondly, the method did not mention how to deal with 2-population \textit{multiplayer} asymmetric games.
Specifically, the experiments for asymmetric game in \cite{tuyls2018generalised} are based on 1 versus 1 games.
To the best of our knowledge, no existing work explicitly describes the methodology of analysing asymmetric games in the multiplayer case.


In this paper, we improve the state-of-the-art method \cite{bloembergen2015evolutionary,tuyls2018generalised} and fill the gaps between a heuristic payoff table and the corresponding evolutionary dynamics without any inaccuracy. In addition, our method provides an accurate mathematical description not only for symmetric games, but also for asymmetric games with multiplayers. This is achieved by taking strategy profiles from both populations into account to develop the generalised formulas for asymmetric HPT. Furthermore, we propose a general framework for $m$ versus $n$ 2-population multiplayer asymmetric games. Subsequently, we compare our method with the state-of-the-art numerically in two classic games: the Prisoner's Dilemma and the Battle of Sexes \cite{gibbons1992game}. Finally, we illustrate the advantages of our method experimentally in two popular domains, Wolfpack and StarCraft II, which both involve complex multiagent interactions. In both experiments, we perform empirical game-theoretical analysis to make the interactions game-theoretically tractable, then utilize multiagent deep reinforcement learning to generate a payoff table for the abstracted game, in order to analyse the evolutionary dynamics.

The rest of paper is organized as follows. Section 2 provides the preliminaries for readers to review some basic concepts. Section 3 introduces our method for both symmetric game and multiplayer asymmetric game. In section 4, two standard examples in game theory are given to compare our method with a state-of-the-art method. In section 5, empirical results are provided in two popular domains, Wolfpack and StarCraft II. Lastly, section 6 presents the conclusion.

\section{Preliminaries}
\subsection{Normal Form Games and Nash Equilibrium}
\subsubsection{Normal Form Games}
A Normal form game (NFG) captures the strategic interaction of players in a single round \cite{morgenstern1944theory,gibbons1992game}. An example of $2$-player NFG is shown in Table \ref{table:2pNFG}.
\begin{table}[t]
\centering
\begin{tabular}{|c|c|c|c|}
\hline
\diagbox{Player 1}{Player 2}&$B_1$&$\cdots$&$B_k$\\
\hline
$A_1$&$(a_{11},b_{11})$&$\cdots$&$(a_{1k},b_{1k})$\\
\hline
$\cdots$&$\cdots$&$\cdots$&$\cdots$\\
\hline
$A_j$&$(a_{j1},b_{j1})$&$\cdots$&$(a_{jk},b_{jk})$\\
\hline
\end{tabular}
\caption{Payoff table for 2-player NFG}\label{table:2pNFG}
\end{table}
Given the strategy profile $(\textbf{x},\textbf{y})$ and payoff bimatrix $(U^1,U^2)$, the corresponding payoff functions for player $1$ and $2$ are $\textbf{x}^T U^1 \textbf{y}$ and $\textbf{x}^T U^2 \textbf{y}$ respectively. Note that these can be generalised into $n$-player case by considering $U^i$ as higher order tensors instead of matrices.
\subsubsection{Nash Equilibrium}
A Nash equilibrium (NE) is an outcome in which no player can get higher payoff if unilaterally changes its strategy, i.e., everyone but itself keeps the strategy as it was. The concept of NE has a number of potential weaknesses. The most glaring weakness is that the concept relies on the strong assumption of hyper-rationality with infinite computational power and complete information of the game. Secondly, the NE only takes account for the equilibrium, a result without the knowledge of how the players proceed. Lastly but not the least, there may exist many Nash Equilibria, and deciding which one should be considered as the outcome of the game is challenging.

\subsection{Symmetric Games and Heuristic Payoff Table}
\subsubsection{Symmetric Games}
An n-player NFG is symmetric in the sense that each player has the same pure strategy set, and one's payoff only depends on the composition of strategy profile rather than who is playing it. More specifically, one's payoff only depends on the numbers of players playing each pure strategy. The simple case is when $n=2$, $A=B^T$. 
\subsubsection{Heuristic Payoff Table}
Thanks to the symmetricity, the heuristic payoff table (HPT) can significantly simplify the payoff table without any loss of information. For instance, HPT of a $3$-player $3$-strategy symmetric game is given as Table \ref{table:sHPT}. The left-hand side, each row in the matrix $N$, indicates the discrete distribution of three players over three strategies, and the right side, the matrix $U$, expresses the rewards corresponding to the rows in $N$.
    \begin{table}[ht]
      \centering
      \begin{tabular}{p{0.9cm}<{\centering}p{0.9cm}<{\centering}p{0.9cm}<{\centering}|p{0.9cm}<{\centering}p{0.9cm}<{\centering}p{0.9cm}<{\centering}}
      \hline
      $N_{i1}$ &$N_{i2}$ &$N_{i3}$ &$U_{i1}$ &$U_{i2}$ &$U_{i3}$\\
      \hline
      3 & 0 & 0 & 1 & 0 & 0\\
      2 & 1 & 0 & 0.5 & 0.2 & 0\\
      \ & $\cdots$ & \ & \ & $\cdots$ & \ \\
      1 & 1 & 1 & 0.1 & 0.2 & 0.4\\
      \ & $\cdots$ & \ & \ & $\cdots$ & \ \\
      0 & 0 & 3 & 0 & 0 & 0.9\\
      \hline
       \end{tabular}
       \caption{HPT for a symmetric game}\label{table:sHPT}
    \end{table}\newline

\subsection{Replicator Dynamics and Evolutionary Stable Strategies}
\subsubsection{Replicator Dynamics}
Evolutionary game theory (EGT) is the study of the dynamic evolution of strategic behaviours. EGT is descriptive comparing with classical game theory, i.e., it captures the complete dynamics of strategies of players. EGT relaxes the rationality assumption and replaces it by biological settings. The essence of EGT is the replicator dynamics (RD), a system of ordinary differential equations (ODEs) that describe how populations of individuals evolve over time under evolutionary pressure. The basic form of RD modelling the selection mechanism for single population is as follows:
\begin{equation}\label{eq:RD1}
\frac{dx_i}{dt}=\left[ (A\textbf{x})_i-\textbf{x}^TA\textbf{x}\right] \cdot x_i.
\end{equation}
Here we consider $k$ social types indexed via $i$ in the population. Each ODE describes the dynamics of the density of a certain type (or pure strategy) in the population. $\textbf{x}=(x_1,\cdots,x_k)$ is a profile (or mixed strategy) of the population that describes the population share of each type at time $t$. $A$ is payoff table so the $(A\textbf{x})_i$ denotes the payoff which the individual receives if choose type $i$ in a population with profile $\textbf{x}$ and $\textbf{x}^TA\textbf{x}$ is the average payoff in the population. The RD models the tendency of types with greater than average payoff to attract more followers (replicators), and types with less than average payoff to suffer defections.

The RD assumes a constant population, i.e., $\sum\limits_i x_i = 1$, then models the rate of change of a given type $i$ be proportional to its current density and its payoff compared to the average payoff of the population. In multiagent setting, an alternative understanding for the evolutionary dynamics is that the agent (individual) learns to choose action (type) given the strategy profile $\textbf{x}$ (population profile) \cite{smith1973logic}.

Similarly, RD for 2-population is described by the system:
\begin{equation}\label{eq:RD2}
\left\{
\begin{array}{l}
dx_i/dt=\left[ (A \textbf{y})_i-\textbf{x}^T A \textbf{y}\right]\cdot x_i\\
dy_j/dt=\left[ (\textbf{x}^TB)_j-\textbf{x}^T B \textbf{y}\right]\cdot y_j,
\end{array}
\right.
\end{equation}
with the matrices $A$ and $B$ being the payoff tables, $\textbf{x}$ and $\textbf{y}$ being the population shares for the first and second population respectively. Note that when the two populations are homogeneous, the game becomes symmetric. Hence $A=B^T$ and the equations (\ref{eq:RD2}) reduce to equations (\ref{eq:RD1}) in single population case.

\subsubsection{Evolutionary Stable Strategies}
The concept of the evolutionary stable strategy (ESS) plays an important role in EGT which can be considered as an analogy to the NE in classical game theory. The ESS is a refined subset of NE. It captures the notion of robustness against invasions in natural selection. In a nutshell, a population profile in which all play the ESS cannot be invaded by a mutant strategy present in a small population share.

For a specific game, all the NE are stationary points of the RD. Whilst the reverse statement is not true, all the attractors from stationary points of RD are indeed ESS, hence are also NE. So the RD of specific game can provide us not only the evolution process, but the ESS.

\section{Method}
\label{sec:method}
As mentioned in Section~\ref{sec:intro}, HPT can crucially simplify the payoff table.
In Section \ref{sec:methodSym}, we firstly improve and detail the calculation in order to fill the gaps between RD and a HPT for a symmetric game, then provide an example to compare our method with a state-of-the-art method mentioned in \cite{tuyls2018generalised}. Then, in Section \ref{sec:methodAsym}, we generalise our method to an asymmetric game with $m+n$ players.
\subsection{Symmetric Games with HPT}\label{sec:methodSym}
We firstly consider symmetric games, i.e., games with monomorphic population and interchangeable individuals. In order to perform RD for a mixed strategy profile $\textbf{x}$, we need to estimate the expected payoff of choosing $i$-th pure strategy, which is the relative Darwinian fitness of choosing a certain type depending on the current population profile in evolutionary game theoretic terms. Had a classical payoff table $A$ in NFG form in hand, this is exactly $(A\textbf{x})_i$ (or $e_iA\textbf{x}$). When provided a HPT, we derive the expected payoff function $f_i(\textbf{x})$ as follows.

Consider an $n$-player $k$-strategy symmetric game with a HPT: $H=\left((N_{ij}),(U_{ij})\right)$. Although there are only finite players interacting in the iterated game, we assume that in each game the $n$ players are randomly drawn from an infinite population pool. This assumption assures that the arbitrary mixed strategy profile and its continuous dynamics make sense. Given a strategy profile $\textbf{x}=(x_1,\cdots ,x_k)$, a variant of combination number $\prescript{k}{n}{\mathcal{N}_j(i)}$ is introduced to denote the combination of the selected $n$ players with the specific row $N_j$ as current population profile if one chooses the $i$-th pure strategy against others. Formally, $\prescript{k}{n}{\mathcal{N}_j(i)}$ is defined as:

\begin{align*}
\prescript{k}{n}{\mathcal{N}}_j(i) &=\binom{n-1}{N_{j1},\cdots,N_{j,i-1},N_{ji}-1,N_{j,i+1},\cdots,N_{jk}}\\
&={n-1 \choose N_{j1}}\cdot \binom{n-1-N_{j1}}{N_{j2}}\cdots \binom{n-1-\sum\limits_{l=1}^{i-2}N_{jl}}{N_{j,i-1}}\\
&\cdot \binom{n-1-\sum\limits_{l=1}^{i-1}N_{jl}}{N_{j,i}-1}\cdot \binom{n-\sum\limits_{l=1}^i N_{jl}}{N_{j,i+1}}\cdot \\
&\cdots \binom{n-\sum\limits_{l=1}^{k-1}N_{jl}}{N_{j,k}}.
\end{align*}
Then the probability for getting a specific row $N_j$ provided the one we consider chooses the $i$-th pure strategy can be computed from:

\begin{equation*}
P(N_j|\textbf{x},i)=
\begin{cases}
\prescript{k}{n}{\mathcal{N}}_j(i)\cdot \left(\prod\limits_l x_l^{N_{jl}}\right)/x_i,& N_{ji}\neq 0,\\
0,& N_{ji}=0.
\end{cases}
\end{equation*}
Note that when one chooses the $i$-th strategy, in the profile the $N_{ji}$ is at least $1$, i.e., it is impossible to get the row $N_j$ with $N_{ji}=0$. Finally, the expected payoff function $f_i(\textbf{x})$ of mixed strategy profile $\textbf{x}$ conditioned on that the one choosing the $i$-th strategy, is computed as the weighted combination of the payoffs given in all rows:

\begin{equation*}
f_i(\textbf{x})=\sum_j P(N_j|\textbf{x},i)\cdot U_{ji}.
\end{equation*}
Then one can substitute $(A\textbf{x})_i$ by $f_i(\textbf{x})$ in RD equations (\ref{eq:RD1}) to analyse the dynamics of strategy profile.
\subsection{Asymmetric Games with HPT}
\label{sec:methodAsym}
In this section, we consider asymmetric games with two population and $m+n$ players, in which the $m$ players are in one population and the other $n$ players are in the other population. In other words, the $m$ players and the $n$ players are homogeneous in their own population respectively. Each player $p\in \{1,\cdots,m+n\}$ in the game can choose a pure strategy $f$ from its meta-strategy set. Note that we can always add dominated strategies to the players who have smaller size of meta-strategy set. Hence without loss of generality, assume that the players in the two populations have a same size $k$ of the meta-strategy sets. More specifically, each of the $m$ players from the first population can choose its own pure strategy from the meta-strategy set $S^{(1)}=\{S^{(1)}_1,\cdots,S^{(1)}_k\}$ and each the $n$ players from the second population can choose a pure strategy from the meta-strategy set $S^{(2)}=\{S^{(2)}_1,\cdots,S^{(2)}_k\}$. The HPT $H=\left((N_{ij}),(U_{ij})\right)$ for the $(m+n)$-player asymmetric game is given as Table \ref{table:mnHPT}. Each entry $(N_{ij}^{(1)},N_{ij}^{(2)})$ of the left part or $(U_{ij}^{(1)},U_{ij}^{(2)})$ of the right part is a $2$-tuple, in which the first and second elements correspond to the counts or rewards information for players from the first and the second populations, respectively.
    \begin{table}[ht]
      \scriptsize
      \centering
      \begin{tabular}{p{0.8 cm}<{\centering}p{0.2 cm}<{\centering}p{0.1 cm}<{\centering}p{0.4 cm}<{\centering}| p{1.3 cm}<{\centering}p{1 cm}<{\centering}p{0.1 cm}<{\centering}p{1.1 cm}<{\centering}}
      \hline
      $N_{i1}$ &$N_{i2}$ & $\cdots$ &$N_{ik}$ &$U_{i1}$ &$U_{i2}$ & $\cdots$ &$U_{ik}$\\
      \hline
      (m,n) & 0 & $\cdots$ & 0 & $(U^{(1)}_{11},U^{(2)}_{11})$ & 0 & $\cdots$ & 0\\
      (m-1,n) & (1,0) & $\cdots$ & 0 & $(U^{(1)}_{21},U^{(2)}_{21})$ & $(U^{(1)}_{22},0)$ & $\cdots$ & 0\\
      (m,n-1) & (0,1) & $\cdots$ & 0 & $(U^{(1)}_{31},U^{(2)}_{31})$ & $(0,U^{(2)}_{32})$ & $\cdots$ & 0\\
      $\vdots$ & $\vdots$ & $\ddots$ & $\vdots$& $\vdots$ & $\vdots$ & $\ddots$ & $\vdots$ \\
      0 & (m,n) & $\cdots$ & 0 & 0 & $(U^{(1)}_{j2},U^{(2)}_{j2})$ & $\cdots$ & 0\\
      $\vdots$ & $\vdots$ & $\ddots$ & $\vdots$& $\vdots$ & $\vdots$ & $\ddots$ & $\vdots$ \\
      0 & 0 & $\cdots$ & (m,n) & 0 & 0 & $\cdots$ & $(U^{(1)}_{rk},U^{(2)}_{rk})$\\
      \hline
       \end{tabular}
       \caption{HPT for an asymmetric game}\label{table:mnHPT}
    \end{table}\newline
The table has $r = \binom{m+k-1}{m} \cdot \binom{n+k-1}{n}$ rows. Once we get the HPT in hand, we can analyse the dynamics of population shares (or strategy distributions) changing in time as following.

Similar to the symmetric case, we need to estimate the expected payoff function $\textbf{f}_i(\textbf{x},\textbf{y})$ for players choosing the $i$-th strategy in both populations in the asymmetric game. For a current joint strategy profile $(\textbf{x},\textbf{y})$, the probability for getting a specific row $N_j$ conditioned on the one we consider from either population chooses the $i$-th strategy can be computed as
\begin{align*}
&P^{(1)}(N_j|\textbf{x},\textbf{y},i)\\
=&\prescript{k}{m}{\mathcal{N}_j^{(1)}(i)}\cdot \left( \prod\limits_{l=1}^k x_l^{N_{jl}^{(1)}}\right)/x_i \\
\cdot &\binom{n}{N_{j1}^{(2)},\cdots ,N_{jk}^{(2)}}\cdot \left( \prod\limits_{l=1}^k y_l^{N_{jl}^{(2)}}\right)\\
=&\prescript{k}{m}{\mathcal{N}_j^{(1)}(i)} \cdot \binom{n}{N_{j1}^{(2)},\cdots ,N_{jk}^{(2)}} \cdot \left( \prod\limits_{l=1}^k x_l^{N_{jl}^{(1)}}y_l^{N_{jl}^{(2)}}\right) /x_i,
\end{align*}
and
\begin{align*}
&P^{(2)}(N_j|\textbf{x},\textbf{y},i)\\
=&\binom{m}{N_{j1}^{(1)},\cdots ,N_{jk}^{(1)}}\cdot \left( \prod\limits_{l=1}^k x_l^{N_{jl}^{(1)}}\right)\\
\cdot &\prescript{k}{n}{\mathcal{N}_j^{(2)}(i)}\cdot \left( \prod\limits_{l=1}^k y_l^{N_{jl}^{(2)}}\right)/y_i \\
=&\prescript{k}{n}{\mathcal{N}_j^{(2)}(i)} \cdot \binom{m}{N_{j1}^{(1)},\cdots ,N_{jk}^{(1)}} \cdot \left( \prod\limits_{l=1}^k x_l^{N_{jl}^{(1)}}y_l^{N_{jl}^{(2)}}\right) /y_i,
\end{align*}
where
$$\prescript{k}{m}{\mathcal{N}_j^{(1)}(i)}=\binom{m-1}{N_{j1}^{(1)},\cdots ,N_{j,i-1}^{(1)},N_{ji}^{(1)}-1,N_{j,i+1}^{(1)},\cdots ,N_{jk}^{(1)}},$$
and
$$\prescript{k}{n}{\mathcal{N}_j^{(2)}(i)}=\binom{n-1}{N_{j1}^{(2)},\cdots ,N_{j,i-1}^{(2)},N_{ji}^{(2)}-1,N_{j,i+1}^{(2)},\cdots ,N_{jk}^{(2)}}.$$
Hence players from either population who choose $i$-th strategy can get corresponding expected payoff
$$f_i^{(1)}(\textbf{x},\textbf{y})=\sum_j P^{(1)}(N_j|\textbf{x},\textbf{y},i)\cdot U_{ji}^{(1)},$$
and
$$f_i^{(2)}(\textbf{x},\textbf{y})=\sum_j P^{(2)}(N_j|\textbf{x},\textbf{y},i)\cdot U_{ji}^{(2)}.$$
Then one can analyse the dynamics of strategy profile via RD equations (\ref{eq:RD2}) by replacing $(Ay)_i$ and $(x^TB)_j$ by $f_i^{(1)}(\textbf{x},\textbf{y})$ and $f_i^{(2)}(\textbf{x},\textbf{y})$, respectively.

\section{Numerical Results}\label{sec:numResult}
In this section, we give two standard examples in game theory to compare our method with a state-of-the-art method \cite{tuyls2018generalised}. 
\subsection{Prisoner's Dilemma (PD)}
A normal form payoff table for PD is shown as follows.
\begin{table}[!htbp]
\centering
\begin{tabular}{|c|c|c|}
\hline
\diagbox{Prisoner 1}{Prisoner 2}&Cooperate&Defect\\
\hline
Cooperate&$(3,3)$&$(0,5)$\\
\hline
Defect&$(5,0)$&$(1,1)$\\
\hline
\end{tabular}
\caption{Normal form payoff table for PD}\label{table:PDNFG}
\end{table}\newline
This is a symmetric game, the payoff matrix for either prisoner is
$$A=\left(
\begin{matrix}
    3 & 0 \\
    5 & 1
\end{matrix}
\right).$$
Assuming the current strategy profile for either prisoner is $\textbf{x}=(0.5,0.5)^T$, the expected payoff conditioned on that one player chooses the $i$-th strategy ($i\in \{1,2\}$) is
\begin{equation}
\label{eq:fNFG}
(A\textbf{x})_i=
\left(\left(
\begin{matrix}
    3 & 0 \\
    5 & 1
\end{matrix}
\right)
\cdot
\left(
\begin{matrix}
    0.5 \\
    0.5
\end{matrix}
\right)\right)_i
=
\left(
\begin{matrix}
    1.5 \\
    3
\end{matrix}
\right)_i.
\end{equation}
Now assume that we only know a priori the corresponding HPT:
    \begin{table}[ht]
      \centering
      \begin{tabular}{p{1cm}<{\centering}p{1cm}<{\centering}|p{1cm}<{\centering}p{1cm}<{\centering}}
      \hline
      $N_{i1}$ &$N_{i2}$ &$U_{i1}$ &$U_{i2}$\\
      \hline
      2 & 0 & 3 & 0\\
      1 & 1  & 0 & 5\\
      0 & 2 & 0 & 1\\
      \hline
       \end{tabular}
       \caption{HPT for PD}\label{table:PDHPT}
    \end{table}\newline
The fact that HPT simplifies normal form payoff table via symmetricities without losing any information indicates that we shall get a same expected payoff function as it is in NFG. Nevertheless, this is not the case when we use the state-of-the-art formulas. As attached in Appendix, the state-of-the-art method yields
\begin{equation*}
f_i(\textbf{x})=\left(
\begin{matrix}
    1 \\
    \frac{11}{3}
\end{matrix}
\right)_i.
\end{equation*}
There is an error when compared with the original expected payoff in expression (\ref{eq:fNFG}).

Now apply our method to the HPT for the PD. According to the formulas in Section \ref{sec:methodSym}, we can get $\prescript{k}{n}{\mathcal{N}_j(i)}$ and $P(N_j|\textbf{x},i)$ for $n=k=2$, $j\in \{1,2,3\}$ and $i\in \{1,2\}$, respectively. Having those terms in hand we can get
\begin{equation*}
f_i(\textbf{x})=\left(
\begin{matrix}
    1.5 \\
    3
\end{matrix}
\right)_i.
\end{equation*}
which corresponds to the expected payoffs derived from the normal form payoff table (more details in Appendix).
\subsection{The Battle of Sexes (BoS)}
A normal form payoff table for BoS is shown as Table \ref{table:BoSNFG}.
\begin{table}
\centering
\begin{tabular}{|c|c|c|}
\hline
\diagbox{Wife}{Husband}&Opera&Football\\
\hline
Opera&$(3,2)$&$(0,0)$\\
\hline
Football&$(0,0)$&$(2,3)$\\
\hline
\end{tabular}
\caption{Normal form payoff table for BoS}\label{table:BoSNFG}
\end{table}\newline
This is an asymmetric game, the payoff matrices for wife and husband are
$$A=\left(
\begin{matrix}
    3 & 0 \\
    0 & 2
\end{matrix}
\right),\ \ and\ \ B=\left(
\begin{matrix}
    2 & 0 \\
    0 & 3
\end{matrix}
\right),$$
respectively. Setting the initial strategy profile be $\textbf{x}=\textbf{y}=(0.5,0.5)^T$, we can easily get the expected payoffs:
\begin{equation}
\label{eq:fNFG2}
(A\textbf{y})_i=
\left(
\begin{matrix}
    1.5 \\
    1
\end{matrix}
\right)_i\ \ and\ \  
(\textbf{x}^TB)_j=
\left(
\begin{matrix}
    1 \\
    1.5
\end{matrix}
\right)_i,
\end{equation}
Now assume that we only have the corresponding HPT in Table \ref{table:BoSHPT}.
\begin{table}
      \centering
      \begin{tabular}{p{1cm}<{\centering}p{1cm}<{\centering}|p{1cm}<{\centering}p{1cm}<{\centering}}
      \hline
      $N_{i1}$ &$N_{i2}$ &$U_{i1}$ &$U_{i2}$\\
      \hline
      (1,1) & 0 & (3,2) & 0\\
      (1,0) & (0,1)  & 0 & 0\\
      (0,1) & (1,0) & 0 & 0\\
      0 & (1,1) & 0 & (2,3)\\
      \hline
       \end{tabular}
       \caption{HPT for BoS}\label{table:BoSHPT}
\end{table}\newline
According to the method claimed by Tuyls, \textit{et al.} in 2018, one should get two decomposed HPT for the wife and for the husband. Then compute expected payoff for either player. For Table \ref{table:HPT1}, one gets the expected payoff for wife:
\begin{equation*}
f_i(\textbf{y})=
\left(
\begin{matrix}
1\\
\frac{2}{3}
\end{matrix}
\right)_i\ \ and\ \ 
f_i(\textbf{x})=
\left(
\begin{matrix}
\frac{2}{3}\\
1
\end{matrix}
\right)_i.
\end{equation*}
Again, there is an error when compared with the original expected payoff in expression (\ref{eq:fNFG2}).

Now apply our method to the HPT for BoS. Formulas in Section \ref{sec:methodAsym} yield

\begin{equation*}
f^{(1)}(\textbf{x},\textbf{y})=
\left(
\begin{matrix}
1.5\\
1
\end{matrix}
\right)_i\ \ and\ \ 
f^{(2)}(\textbf{x},\textbf{y})=
\left(
\begin{matrix}
1\\
1.5
\end{matrix}
\right)_i.
\end{equation*}
which correspond to the expected payoffs (\ref{eq:fNFG2}) derived from the normal form payoff table. For more details of the computations, please see the Appendix section.

Both PD and BoS games indicate that our method can derive an accurate expected payoff from HPT without losing any information from original NFG, in order to perform RD to study the evolutionary dynamics.


\begin{figure}
  \begin{subfigure}[b]{0.23\textwidth}
  \includegraphics[width=\textwidth]{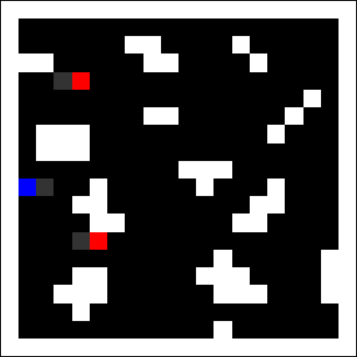}
  \caption{The Wolfpack environment}
  \label{fig:wp_env}
  \end{subfigure}
  \begin{subfigure}[b]{0.23\textwidth}
  \includegraphics[width=\textwidth]{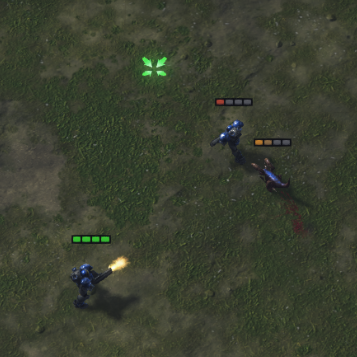}
  \caption{The StarCraft II environment}
  \label{fig:sc2_env}
  \end{subfigure}
  \caption{Complex environments}\label{fig:envs}
\end{figure}

\section{Empirical Results}
In this section, we conduct two experiments based on the empirical game-theoretic analysis (henceforth EGTA, Walsh 2002, Wellman 2006) which focuses on so called meta-strategies in a high-level strategic view, rather than atomic actions. Basically, a potentially very complex, multi-stage game is reduced to a one-shot NFG, which is tractable to study the dynamics analytically.
\subsection{Wolfpack Domain}
The wolfpack game \cite{leibo2017multi} involves three players, two wolves as the predators and a sheep as the prey. Instead of regarding both predators and prey as players, we consider the sheep, who adopts the random walk as strategy, as part of the environment, and treat the two wolves as two different populations. 
Then, the capture in the grid world is defined as whenever one of wolves appears in the adjacent grid of the sheep, and the wolf will receive a basic reward for grabbing the sheep, defined as $r_{lone}$. 
When a capture occurs, the distances, i.e., Manhattan distance between wolves and sheep are less than a pre-defined threshold, then, the two wolves acquire extra bonus, $r_{team}$, respectively.
The basic idea is that the wolves would cooperate to share the prey in order to have a better and faster meal. However, if the capture happens, with a wolf too far away from the prey, the one who seizes the sheep would only get the basic reward, $r_{lone}$, and the reward of the other one is zero.

We select two meta-strategies for the two wolves, trained via Proximal Policy Optimization (PPO) \cite{schulman2017proximal}, in different environment settings. Each wolf is modeled with recurrent neural network with its own parameters. With high/low $r_{team}$ for wolves, the trained strategies are classified as cooperative(C)/defective(D). The two wolves share the same strategy set, but different sampling policies. We consider the two wolves as two populations, and denote the set of cooperative, defective policies of different wolves as $\Pi^C$, $\Pi^D$ respectively. 

We stochastically choose two policy pairs $(\pi ^C_1,\pi ^D_1)$ and $(\pi ^C_2,\pi ^D_2)$ to simulate the game in one episode, where $\pi _{1,2}^C\in \Pi^C$ and $\pi _{1,2}^D\in \Pi^D$, then update the reward in the following HPT:
    \begin{table}[ht]
      \centering
      \begin{tabular}{p{1cm}<{\centering}p{1cm}<{\centering}|p{2cm}<{\centering}p{2cm}<{\centering}}
      \hline
      $N_{i1}$ &$N_{i2}$ &$U_{i1}$ &$U_{i2}$\\
      \hline
      (1, 1) & 0 & (1.32, 1.34) & 0 \\ 
      (1, 0) & (0, 1) & (0.82, 0) & (0, 1.53) \\ 
      (0, 1) & (1, 0) & (0, 0.81) & (1.53,0) \\ 
      0 & (1, 1) & 0 & (0.74, 0.72) \\
      \hline
      \end{tabular}
      \caption{HPT for Wolfpack}\label{table:HPW}
    \end{table}\newline
We repeat sampling, simulating and updating in independent episodes as many as possible until the HPT converges.

\begin{figure}[t]
\centering
\includegraphics[width = 8cm ]{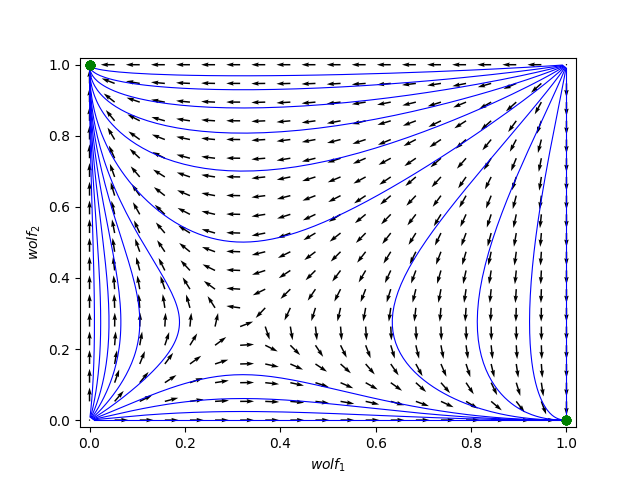}
\caption{Evolutionary dynamics for the Wolfpack experiment via our method}\label{fig:WPU}
\vspace{-10pt}
\end{figure}

\begin{figure}[t]
\centering
\includegraphics[width = 8cm ]{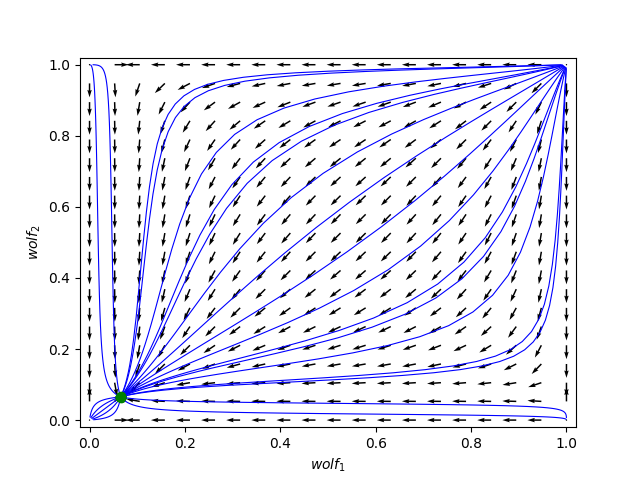}
\caption{Evolutionary dynamics for the Wolfpack experiment via the state-of-the-art method}\label{fig:WPDM}
\vspace{-10pt}
\end{figure}

In Figure \ref{fig:WPU}, the evolutionary dynamics of the meta-game payoff table is illustrated, i.e., directional field and trajectory plot, using the replicator dynamics described in Equation \ref{eq:RD2}. With x-axis as wolf$_1$ and y-axis as wolf$_2$, the arrows indicate the direction of strategies evolution, and the convergence point of blue curves is labeled by green circle. The coordinate is directly proportional to the degree of cooperation, so the point $[0.0, 1.0]$ represents to fully defective strategies $\Pi^D$ and fully cooperative strategies $\Pi^C$. The situation is a bit different from the Figure \ref{fig:WPDM} drawn by replicator dynamics equation proposed in \cite{tuyls2018generalised}. From the perspective of team, the pairs of strategies is sorted by wolves’ total payoff as $(\pi_1^C, \pi_2^C) > (\pi_1^C, \pi_2^D) = (\pi_1^D, \pi_2^C) > (\pi_1^D, \pi_2^D)$. Here we observe three equilibria in the direction field showed in Figure \ref{fig:WPU}: one pure at $(0.0, 1.0)$, one pure at $(1.0, 0.0)$, and one unstable mixed equilibrium at the point $(0.32, 0.28)$ which is insignificant in our current analysis. The first two equilibria, also called attractors, presents $(\pi_1^C, \pi_2^D)$ and $(\pi_1^D, \pi_2^C)$ respectively. However, the equilibrium at $(0.07, 0.07)$ in Figure \ref{fig:WPDM} approximates $(\pi_1^D, \pi_2^D)$, which indicates the two wolves intend to choose competition instead of cooperation. Our dynamics result shows that at least one wolf would choose to cooperate, which is consistent with the actual behaviors we observed in the visual interface.

To illustrate, we provide a detailed description of strategies evolution. The one who takes cooperative strategy will wait for another wolf to approach to the sheep. It is reasonable that the wolves both receive relatively high rewards when the pair of strategies is $(\pi_1^C, \pi_2^C)$. It is worthwhile to mention that, when the team breaks up because of one wolf's betraying, e.g., the pair shifting from $(\pi_1^C, \pi_2^C)$ to $(\pi_1^C, \pi_2^D)$, the one who takes the defective strategy will get higher rewards than the cooperative situation. The cooperative wolf will choose to wait for another when the prey appears within its vision, while the defective one rushes to the prey directly no matter where the teammate is.

\subsection{StarCraft II Domain}
Peng peng et al. proposed BiCNet to study the cooperative behavior among three marines when facing their common enemy, a zergling, using the StarCraft as a test bed. They observed that the agents can learn various coordinating strategies under different environmental settings. Especially when the marines play in a tough environment, self-sacrifice would happen among them to win the game \cite{peng2017multiagent}. In our experiment, we reduce the number of marines into two so as to improve the possibility of revealing the cooperative relationship between marines.
Consider the marines and zergling as two populations.
The goal of each population is to kill the unit(s) in another population with minimum lost,
with rewards calculated as their personal damage to the enemy subtracts their own health points (HP) lost. Each marine has 45 HP and the zergling has 250. Both can produce 5 HP damage to the opponent per attack, with different attacking styles: a marine takes a shot in range while the zergling attacks the enemy when they are close.
Similar to the Wolfpack domain, we fixed the strategy of the zergling, and train two marines via Proximal Policy Optimization (PPO)~\cite{schulman2017proximal}. Two marines are modeled using Recurrent Neural Network with sharing parameters. Next, we firstly describe the strategies of the zergling, and then describe the corresponding strategies of marines learned accordingly.

The strategies of the zergling depend on its territory size,
which corresponds to the zergling's degree of fierceness.
We setup two experiments in terms of different territory sizes, one of which covers the whole map of size $14$ by $14$, and another of which is round in shape with radius equal to $6$. The first setup corresponds to the aggressive strategy of the zergling ($A$) for which the zergling attacks throughout the map,
while the second setup is referred to the passive strategy ($P$),
for which the zergling attacks only inside the territory.


In the first experiment, we observe that two marines learn to cooperate with each other: one marine draws attention from the zergling and run away smartly in the map, while another marine will focus fire on the enemy. We refer to the policy learned as the cooperative strategy ($C$).
In the second experiment, we observe that both marines behave similarly, either run away from the zergling or shoot the zergling at the same time. We refer to the policy learned in this setting as the duplicated strategy ($D$).


Slightly different from wolfpack, we denote the cooperative, duplicated, aggressive and passive set of policies as $\Pi^C$, $\Pi^D$, $\Pi^A$ and $\Pi^P$.
As described above, the size of either $\Pi^A$ or $\Pi^P$ is 1. For marines', we re-run the two experiments with various parametric initialization and add the learned policies into the set $\Pi^C$ and $\Pi^D$, respectively.
Then, we sample $\pi_{1,2}^C \in \Pi^C$, $\pi_{1,2}^D \in \Pi^D$, $\pi_a \in \Pi ^A$ and $\pi_p \in \Pi ^P$ to simulate, record and update the reward into the HPT. 

\begin{figure}[t]
\centering
\includegraphics[width = 8cm ]{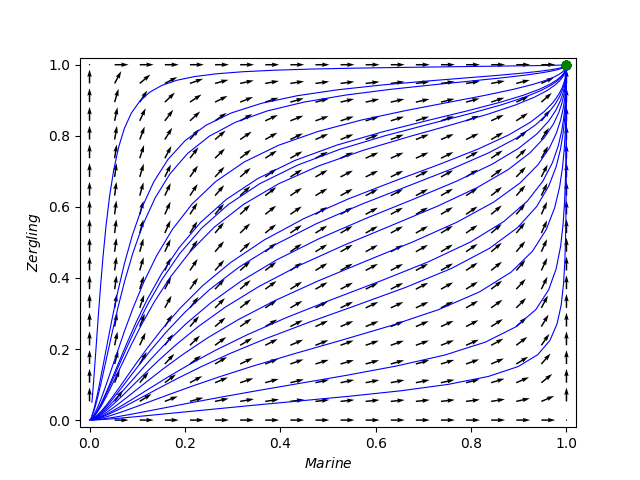}
\caption{Evolutionary dynamics for StarCraft II experiment}\label{fig:SC2}
\vspace{-10pt}
\end{figure}

The HPT for ``2 Marines versus 1 Zergling'' in StarCraft II is shown in the following:
    \begin{table}[ht]
      \centering
      \begin{tabular}{p{1cm}<{\centering}p{1cm}<{\centering}|p{2.2cm}<{\centering}p{2.2cm}<{\centering}}
      \hline
      $N_{i1}$ &$N_{i2}$ &$U_{i1}$ &$U_{i2}$\\
      \hline
      (2, 1) & 0 & (104.5, -209.0) & 0 \\
      (2, 0) & (0, 1) & (117.3, 0) & (0, -234.6) \\
      (1, 1) & (1, 0) & (68.2, -118.9) & (50.7, 0) \\
      (1, 0) & (1, 1) & (93.4, 0) & (56.4, -149.8) \\
      (0, 1) & (2, 0) & (0, -105.4) & (52.7, 0) \\
      0 & (2, 1) & 0 & (70.4, -140.8) \\
      \hline
      \end{tabular}
      \caption{HPT for StarCraft II}\label{table:HPSC2}
      \vspace{-10pt}
    \end{table}\newline
Since ``2 Marines versus 1 Zergling'' is a multiplayer asymmetric game, we cannot use the method proposed in \cite{tuyls2018generalised} for computing the game dynamics. Thus, we only show the directional field and trajectory plot drawn via our method. In Figure~\ref{fig:SC2}, the x-axis represents the probability with which two marines play cooperatively, while the y-axis represents the probability with which the zergling plays aggressively. We observe that all dynamics are absorbed by the equilibrium $(\Pi^A, \Pi^C)$. Combined with the illustration with analysis of the Table \ref{table:HPSC2}, we see that $(\Pi^A, \Pi^C)$ dominates all the other pairs. In addition, we give a semantic description here.

Starting with $(\Pi^P, \Pi^D)$, both sides are incentive to get higher rewards. Without loss of generality, we assume that the zergling first learns the aggressive strategy, keeping running after the marines who are limited to stand to shoot. The marines have not yet learned to cooperate, and most of the time, zergling destroys the enemy one by one. In this case, the reward of zergling is the highest in the Table \ref{table:HPSC2}. Then, one marine changes his strategy to the cooperative strategy for higher rewards, who may run in a round to attract the attention of zergling when the latter is running after him, or stand still to shoot as long as there is a chance. Another marine, however, runs away when chased by zergling and frequently let chance of attacking go. To this marine, a cooperative mate eases the pressure on protecting himself from losing HP, meanwhile, he misses the points of damage, resulting in lower rewards. Finally, both marines adopt cooperative strategies, and the ``division of labor cooperation'' emerges. The one chased by zergling runs in a circle, of which the center his companion stands to shoot at the enemy. It is notable that, from the Figure~\ref{fig:SC2}, we can see that, no matter where the start is, the pair will converge to the equilibrium, $(\Pi^A, \Pi^C)$.

\section{Conclusion}
In this paper, a more accurate method to close the gaps between a heuristic payoff table and the corresponding evolutionary dynamics is introduced. We show that this approach is applicable to both of symmetric and asymmetric games. Moreover, a general framework for $m$ versus $n$ 2-population multiplayer asymmetric games is developed and verified through two classic normal form games. Finally, we empirically demonstrate the advantages of our method in two popular domains, Wolfpack and StarCraft II.

In the future, we plan to extend our framework from $2$-population games into $k$-population games and study the relationship between multiagent reinforcement learning (MARL) and EGT, which may shed lights on the research in both MARL and EGT communities.

\newpage

\bibliography{mybib}
\bibliographystyle{aaai}
\clearpage
\appendix
\section*{Appendix}\label{sec:App}
\subsection*{Computations for Section \ref{sec:numResult}}
\subsubsection{The Prisoner's Dilemma (PD)}
A normal form payoff table for PD is shown in Table \ref{table:PDNFG}. This is a symmetric game, the payoff matrix for either prisoner is
$$A=\left(
\begin{matrix}
    3 & 0 \\
    5 & 1
\end{matrix}
\right).$$
Assuming the current strategy profile for either prisoner is $\textbf{x}=(0.5,0.5)^T$, the expected payoff conditioned on that one player chooses the $i$-th strategy ($i\in \{1,2\}$) is
\begin{equation}
(A\textbf{x})_i=
\left(\left(
\begin{matrix}
    3 & 0 \\
    5 & 1
\end{matrix}
\right)
\cdot
\left(
\begin{matrix}
    0.5 \\
    0.5
\end{matrix}
\right)\right)_i
=
\left(
\begin{matrix}
    1.5 \\
    3
\end{matrix}
\right)_i.
\end{equation}
Now assume that we only know a priori the corresponding HPT as shown in Table \ref{table:PDHPT}. The fact that HPT simplifies normal form payoff table via symmetricities without losing any information indicates that we shall get a same expected payoff function as it is in NFG. Nevertheless, this is not the case when we use the state-of-the-art formulas. The formulas claim that
\begin{equation*}
\left\{
\begin{array}{l}
P(N_j|\textbf{x})=C_{n}^{N_{j1},N_{j2}}\prod\limits_{i=1}^2x_i^{N_{ji}},\\
f_i(\textbf{x})=\frac{\sum\limits_{j}P(N_j|\textbf{x})U_{ji}}{1-(1-x_i)^2}.
\end{array}
\right.
\end{equation*}
Then
\begin{equation*}
\left\{
\begin{array}{l}
P(N_1|\textbf{x})=1\cdot 0.5^2+0=0.25,\\
P(N_2|\textbf{x})=2\cdot 0.5\cdot 0.5=0.5,\\
P(N_3|\textbf{x})=0+1\cdot 0.5^2=0.25.
\end{array}
\right.
\end{equation*}
which yields
\begin{equation*}
\left\{
\begin{array}{l}
f_1(\textbf{x})=\frac{P(N_1|\textbf{x})U_{11}+P(N_2|\textbf{x})U_{21}+P(N_3|\textbf{x})U_{31}}{1-(1-x_1)^2}=1,\\
f_2(\textbf{x})=\frac{P(N_1|\textbf{x})U_{12}+P(N_2|\textbf{x})U_{22}+P(N_3|\textbf{x})U_{32}}{1-(1-x_2)^2}=\frac{11}{3}.
\end{array}
\right.
\end{equation*}
There is an error when compared with the original expected payoff in expression (\ref{eq:fNFG}).

Now apply our method to the HPT for the PD. Formulas in Section \ref{sec:methodSym} state that
\begin{equation*}
\left\{
\begin{array}{l}
P(N_1|\textbf{x},1)=\prescript{2}{2}{\mathcal{N}_1(1)}\cdot 0.5^2/0.5,\\
P(N_1|\textbf{x},2)=0,\\
P(N_2|\textbf{x},1)=\prescript{2}{2}{\mathcal{N}_2(1)}\cdot 0.5\cdot 0.5/0.5,\\
P(N_2|\textbf{x},2)=\prescript{2}{2}{\mathcal{N}_2(2)}\cdot 0.5\cdot 0.5/0.5,\\
P(N_3|\textbf{x},1)=0,\\
P(N_3|\textbf{x},2)=\prescript{2}{2}{\mathcal{N}_3(2)}\cdot 0.5^2/0.5,\\
\end{array}
\right.
\end{equation*}
with
$$
\prescript{2}{2}{\mathcal{N}_1(1)}=
\prescript{2}{2}{\mathcal{N}_2(1)}=
\prescript{2}{2}{\mathcal{N}_2(2)}=
\prescript{2}{2}{\mathcal{N}_3(2)}=\binom{1}{1}=1.
$$
Having above terms in hand we can get
\begin{flalign*}
f_1(\textbf{x})&=P(N_1|\textbf{x},1)\cdot U_{11}+P(N_2|\textbf{x},1)\cdot U_{21}\\
&+P(N_3|\textbf{x},1)\cdot U_{31}=1.5,
\end{flalign*}
and
\begin{flalign*}
f_2(\textbf{x})&=P(N_1|\textbf{x},2)\cdot U_{12}+P(N_2|\textbf{x},2)\cdot U_{22}\\
&+P(N_3|\textbf{x},2)\cdot U_{32}=3,
\end{flalign*}
which correspond to the expected payoffs derived from the normal form payoff table.
\subsubsection{The Battle of Sexes (BoS)}
A normal form payoff table for BoS is shown as Table \ref{table:BoSNFG}. This is an asymmetric game, the payoff matrices for wife and husband are
$$A=\left(
\begin{matrix}
    3 & 0 \\
    0 & 2
\end{matrix}
\right),\ \ and\ \ B=\left(
\begin{matrix}
    2 & 0 \\
    0 & 3
\end{matrix}
\right),$$
respectively. Setting the initial strategy profile be $\textbf{x}=\textbf{y}=(0.5,0.5)^T$, we can easily get the expected payoffs:
\begin{equation}
(A\textbf{y})_i=
\left(
\begin{matrix}
    1.5 \\
    1
\end{matrix}
\right)_i\ \ and\ \  
(\textbf{x}^TB)_j=
\left(
\begin{matrix}
    1 \\
    1.5
\end{matrix}
\right)_j,
\end{equation}
Now assume that we only have the corresponding HPT in Table \ref{table:BoSHPT}. According to the method claimed by Tuyls, \textit{et al.} in 2018, one should get two decomposed HPT:
\begin{table}[!htbp]
      \centering
      \begin{tabular}{p{1cm}<{\centering}p{1cm}<{\centering}|p{1cm}<{\centering}p{1cm}<{\centering}}
      \hline
      $N_{i1}$ &$N_{i2}$ &$U_{i1}$ &$U_{i2}$\\
      \hline
      2 & 0 & 3 & 0\\
      1 & 1 & 0 & 0\\
      0 & 2 & 0 & 2\\
      \hline
       \end{tabular}
       \caption{HPT1}\label{table:HPT1}
\end{table}\newline
for the wife and
\begin{table}[!htbp]
      \centering
      \begin{tabular}{p{1cm}<{\centering}p{1cm}<{\centering}|p{1cm}<{\centering}p{1cm}<{\centering}}
      \hline
      $N_{i1}$ &$N_{i2}$ &$U_{i1}$ &$U_{i2}$\\
      \hline
      2 & 0 & 2 & 0\\
      1 & 1  & 0 & 0\\
      0 & 2 & 0 & 3\\
      \hline
       \end{tabular}
       \caption{HPT2}\label{table:HPT2}
\end{table}\newline
for the husband. Then compute expected payoff for either player. For Table \ref{table:HPT1}, one gets the expected payoff for wife:
\begin{equation*}
\left\{
\begin{array}{l}
f_1(\textbf{y})=\frac{P(N_1|\textbf{y})U_{11}+P(N_2|\textbf{y})U_{21}+P(N_3|\textbf{y})U_{31}}{1-(1-y_1)^2}=1,\\
f_2(\textbf{y})=\frac{P(N_1|\textbf{y})U_{12}+P(N_2|\textbf{y})U_{22}+P(N_3|\textbf{y})U_{32}}{1-(1-y_2)^2}=\frac{2}{3}.
\end{array}
\right.
\end{equation*}
For Table \ref{table:HPT2}, one can get the expected payoff for husband:
\begin{equation*}
\left\{
\begin{array}{l}
f_1(\textbf{x})=\frac{P(N_1|\textbf{x})U_{11}+P(N_2|\textbf{x})U_{21}+P(N_3|\textbf{x})U_{31}}{1-(1-x_1)^2}=\frac{2}{3},\\
f_2(\textbf{x})=\frac{P(N_1|\textbf{x})U_{12}+P(N_2|\textbf{x})U_{22}+P(N_3|\textbf{x})U_{32}}{1-(1-x_2)^2}=1.
\end{array}
\right.
\end{equation*}
Again, there is an error when compared with the original expected payoff in expression (\ref{eq:fNFG2}).

Now apply our method to the HPT for BoS. Formulas in Section \ref{sec:methodAsym} state that
\begin{equation*}
\left\{
\begin{array}{l}
P^{(p)}(N_1|\textbf{x},\textbf{y},1)=\prescript{2}{2}{\mathcal{N}^{(p)}_1(1)}\cdot 0.5^2/0.5,\\
P^{(p)}(N_1|\textbf{x},\textbf{y},2)=0,\\
P^{(p)}(N_2|\textbf{x},\textbf{y},1)=\prescript{2}{2}{\mathcal{N}^{(p)}_2(1)}\cdot 0.5\cdot 0.5/0.5,\\
P^{(p)}(N_2|\textbf{x},\textbf{y},2)=\prescript{2}{2}{\mathcal{N}^{(p)}_2(2)}\cdot 0.5\cdot 0.5/0.5,\\
P^{(p)}(N_3|\textbf{x},\textbf{y},1)=0,\\
P^{(p)}(N_3|\textbf{x},\textbf{y},2)=\prescript{2}{2}{\mathcal{N}^{(p)}_3(2)}\cdot 0.5^2/0.5,\\
\end{array}
\right.
\end{equation*}
with
$$
\prescript{2}{2}{\mathcal{N}^{(p)}_1(1)}=
\prescript{2}{2}{\mathcal{N}^{(p)}_2(1)}=
\prescript{2}{2}{\mathcal{N}^{(p)}_2(2)}=
\prescript{2}{2}{\mathcal{N}^{(p)}_3(2)}=\binom{1}{1}=1,
$$
where $p\in \{1,2\}$. Having above terms in hand we can get
\begin{flalign*}
f^{(1)}_1(\textbf{x},\textbf{y})&=P^{(1)}(N_1|\textbf{x},\textbf{y},1)\cdot U^{(1)}_{11}+P^{(1)}(N_2|\textbf{x},\textbf{y},1)\cdot U^{(1)}_{21}\\
&+P^{(1)}(N_3|\textbf{x},\textbf{y},1)\cdot U^{(1)}_{31}=1.5,\\
f^{(1)}_2(\textbf{x},\textbf{y})&=P^{(1)}(N_1|\textbf{x},\textbf{y},2)\cdot U^{(1)}_{12}+P^{(1)}(N_2|\textbf{x},\textbf{y},2)\cdot U^{(1)}_{22}\\
&+P^{(1)}(N_3|\textbf{x},\textbf{y},2)\cdot U^{(1)}_{32}=1,\\
f^{(2)}_1(\textbf{x},\textbf{y})&=P^{(2)}(N_1|\textbf{x},\textbf{y},1)\cdot U^{(2)}_{11}+P^{(2)}(N_2|\textbf{x},\textbf{y},1)\cdot U^{(2)}_{21}\\
&+P^{(2)}(N_3|\textbf{x},\textbf{y},1)\cdot U^{(2)}_{31}=1,\\
f^{(2)}_2(\textbf{x},\textbf{y})&=P^{(2)}(N_1|\textbf{x},\textbf{y},2)\cdot U^{(2)}_{12}+P^{(2)}(N_2|\textbf{x},\textbf{y},2)\cdot U^{(2)}_{22}\\
&+P^{(2)}(N_3|\textbf{x},\textbf{y},2)\cdot U^{(2)}_{32}=1.5,\\
\end{flalign*}
which correspond to the expected payoffs (\ref{eq:fNFG2}) derived from the normal form payoff table.\\

\end{document}